\documentclass[letterpaper, 10 pt, conference]{ieeeconf}  

\usepackage{fix-cm}

\IEEEoverridecommandlockouts                              

\usepackage{amsmath}
\usepackage{booktabs}
\usepackage{amssymb}
\usepackage{graphicx}
\usepackage[hidelinks]{hyperref}
\title{\LARGE \bf
ACE: Semantic Task Composition for Tabletop Manipulation via Tracked Pick-and-Place Masks
}

\author{
Iok Tong Lei\textsuperscript{1,\ensuremath{\dagger}},
Qianzhi Li\textsuperscript{2,\ensuremath{\dagger}},
Ying Jie Yap\textsuperscript{1},
Yujie Zhang\textsuperscript{5},
Rui Zhong\textsuperscript{3},\\
Haichao Gui\textsuperscript{3,*},
Xiaolong Liu\textsuperscript{4,*},
Zhidong Deng\textsuperscript{1,*}
\thanks{
\textsuperscript{1}
Iok Tong Lei, Ying Jie Yap, and Zhidong Deng are with the Department of Computer Science and Technology, Tsinghua University, Beijing, China.
}
\thanks{
\textsuperscript{2}
Qianzhi Li is with the National College for Excellent Engineers, Beihang University, Beijing, China.
}
\thanks{
\textsuperscript{3}
Rui Zhong and Haichao Gui are with the School of Astronautics, Beihang University, Beijing, China.
}
\thanks{
\textsuperscript{4}
Xiaolong Liu is with Wuxi Dexteroushands Robotic Technology Co.
}
\thanks{
\textsuperscript{5}
Yujie Zhang is an independent researcher.
}
\thanks{
\textsuperscript{\ensuremath{\dagger}}
Iok Tong Lei and Qianzhi Li contributed equally to this work.
}
\thanks{
\textsuperscript{*}
Corresponding authors: Haichao Gui, Xiaolong Liu, and Zhidong Deng.
Emails: \texttt{xllau@126.com}, \texttt{michael@mail.tsinghua.edu.cn}.
Primary contact: Iok Tong Lei, \texttt{lixt22@mails.tsinghua.edu.cn}.
}
}

\begin{document}

\maketitle
\thispagestyle{empty}
\pagestyle{empty}

\begin{abstract}

General-purpose manipulation requires both semantic reasoning over task constraints and reliable execution of contact-rich actions. We present ACE, an agentic manipulation harness that composes a high-level language agent with a reusable mask-conditioned visuomotor policy. Given an open-ended instruction, the agent solves semantic constraints, binds objects to destination roles, and decomposes the task into executable transfers represented by tracked pick-and-place masks. Execution feedback supports outcome assessment, re-grounding, and retry, while persistent object and task context preserves earlier associations when manipulation changes visible cues. We evaluate ACE on two physical multi-step tabletop tasks, Semantic Formula Assembly and Constraint Retrieval. The visuomotor policy is trained only on generic pick-and-place demonstrations and reused without complete demonstrations of either evaluation task, enabling task-level zero-shot composition. Across 20 randomized trials per task, ACE achieves 70\% and 80\% success, respectively, compared with 55\% and 70\% without persistent context. These results suggest that an agentic harness can extend a primitive-trained manipulation policy to semantically distinct tasks through explicit object-destination interfaces and closed-loop execution feedback.

\end{abstract}

\section{Introduction}
\begin{figure*}[t]
    \centering
    \includegraphics[width=1\linewidth]{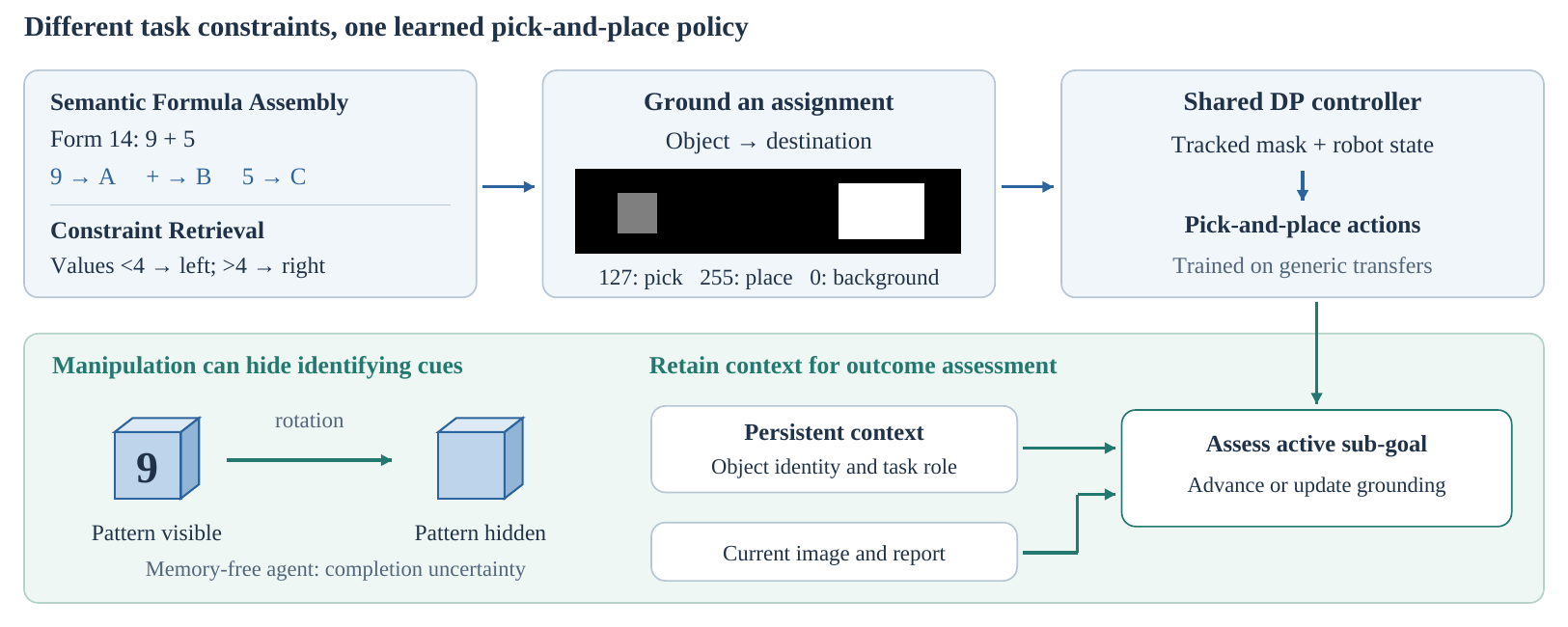} 
     \caption{\textbf{Semantic task composition in ACE.} Different task constraints produce object--destination assignments that share a tracked mask interface and a learned pick-and-place policy. Persistent context retains earlier object and task information for assessing outcomes when manipulation changes visible identifying cues. Masks and cubes are schematic; the full system is shown in Fig.~\ref{fig:framework}.}
    \label{fig:paradigm_comparison}
    \label{fig:contract}
\end{figure*}
Natural language provides an intuitive interface for robotic control, allowing users to specify goals at the task level. Executing these goals requires connecting semantic constraints to physical actions~\cite{shridhar2022cliport,ahn2022doberkeley,zitkovich2023rt}. Instructions may define relationships that several objects must jointly satisfy, alongside the spatial roles those objects must occupy. The robot must resolve these relationships, execute the corresponding transfers, and assess the resulting arrangement.

In Semantic Formula Assembly, the agent selects number and operator cubes that jointly satisfy a requested arithmetic relation, assigns them to operand and operator slots, and checks the expression realized by their placement. Constraint Retrieval maps selection conditions to eligible objects and their destinations. These tasks share a manipulation primitive while requiring different semantic decisions. Formula assembly specifically couples constraint solving, role-to-location binding, and verification of a relation expressed by several physical objects.

Learned manipulation policies provide reusable physical skills~\cite{zeng2018learning,brohan2023rt1}, and language-model agents can compose perception and control calls while incorporating execution feedback~\cite{liang2023codeaspolicies,huang2022inner}. Each semantic decision must identify both an object and its destination, and their image regions must be updated as execution changes the scene. Manipulation can also alter the evidence available for checking completion: a cube may turn so that its identifying pattern is no longer visible. This motivates retaining both current spatial targets and earlier object context.

We present \textbf{ACE} (\textbf{A}gentic \textbf{C}ontrol for \textbf{E}mbodied manipulation), a framework that connects semantic workflows to learned pick-and-place execution through \textbf{unified, tracked object-and-destination masks}. Given an instruction, the agent generates semantic sub-goals and invokes two robot-facing skills: \texttt{MaskInterface} grounds the active sub-goal, and \texttt{PickPlace} executes the corresponding transfer. A single mask assigns distinct pixel values to the pick object and place region. Tracking updates these regions during execution, and the resulting mask conditions a reusable vision-action policy.

The object--destination assignment connects these task requirements to execution: semantic roles specify the intended transfer, and the tracked mask supplies its current image regions. ACE targets \emph{task-level zero-shot composition}. The same controller, trained on generic pick-and-place demonstrations, executes both tasks without full-task demonstrations of either evaluation task. The agent composes object selection, destination assignment, and sequencing at inference (Fig.~\ref{fig:contract}).

ACE separates live workflow and tracking state from persistent object and conversational context. Current masks specify where to act; retained context makes earlier object associations and task information available when assessing the resulting scene. After an attempt, the agent checks the observation and execution report against the active sub-goal, then advances or refreshes grounding. We evaluate this workflow on two tasks and ablate persistent context while keeping the DP controller and live state fixed.

Our contributions are:
\begin{itemize}
    \item A framework connecting task-level constraint solving, object--destination role binding, and learned pick-and-place execution through a shared controller.
    \item An execution interface combining tracked pick/place regions with persistent object and task context for grounding and outcome assessment as manipulation changes the scene.
    \item A physical evaluation achieving 70\% and 80\% success on Semantic Formula Assembly and Constraint Retrieval, with matched visual-input and persistent-context ablations and qualitative analysis of completion uncertainty after cube rotation.
\end{itemize}

\section{Related Work}
\label{sec:related_work}

\subsection{Language-Model Agents for Manipulation}
Large language models (LLMs) and vision-language models (VLMs) support robotic task understanding and embodied planning~\cite{ahn2022doberkeley,huang2022inner,zitkovich2023rt,ma2024surveyvla}, including instruction parsing, task decomposition, and tool use~\cite{driess2023palm-e,yao2022react,qin2024toolllm}. Code as Policies composes perception and control APIs through language-model-generated programs~\cite{liang2023codeaspolicies}. Inner Monologue incorporates environment feedback into task planning~\cite{huang2022inner}. ACE builds on these agent functions and focuses on their connection to a learned pick-and-place primitive: each workflow step supplies an explicit object--destination assignment through a tracked visual mask.

\subsection{Visual Grounding and Mask-Conditioned Control}
Visual grounding and tracking connect task descriptions to targets in changing scenes~\cite{plummer2015flickr30k,kirillov2023segment,cheng2023tracking}. MOO~\cite{pmlr-v229-stone23a} encodes object locations using distinct pixel values and supplies them with RGB observations and a skill instruction to a learned policy. ACE uses tracked object and destination regions to execute semantic workflows with a controller conditioned on masks and proprioception. A closely related system, DexGraspVLA~\cite{zhong2026dexgraspvla}, combines a VLM planner, SAM/Cutie target masks, and a diffusion-based grasping controller. Its planner monitors grasp outcomes and sequences multiple grasping steps; a successful grasp triggers a scripted placing motion in the described pipeline. ACE encodes both the pick object and the place destination in one tracked mask, conditioning a learned policy for the complete pick-and-place transfer. This interface maps semantic placement roles, such as operand and operator slots, to the shared controller.

VoxPoser grounds language into three-dimensional value maps for model-based motion planning~\cite{pmlr-v229-huang23b}. ACE uses a two-dimensional pick-and-place mask as the input to a learned manipulation policy. Object selection and destination assignment are resolved by the agent, while the controller maps the tracked spatial targets and robot state to actions.

\subsection{Task-Level Semantic Requirements}
Task requirements provide a further distinction from these closely related systems. MOO evaluates known manipulation skill templates with novel object descriptions~\cite{pmlr-v229-stone23a}. DexGraspVLA's long-horizon evaluation includes clearing a scene and selecting objects by color or category for sequential grasping~\cite{zhong2026dexgraspvla}. ACE's formula task requires selecting operands and an operator that jointly satisfy an arithmetic relation, binding them to distinct expression slots, and assessing the assembled expression. This links a constraint over several selected objects to role-specific physical placement. Constraint Retrieval shares the condition-based selection structure of such evaluations and assigns selected objects to instruction-specified destinations. Both tasks use the same primitive-trained controller, while ACE composes the workflow at inference.

\subsection{Persistent Context for Sequential Decisions}
Agents use object histories, scene information, and conversational context to retain information across decisions~\cite{park2023generative,wang2023voyager,shinn2024reflexion}. ACE separates live workflow and tracking state from object and conversational context retained across agent calls. The persistent context is available when physical actions change the visual cues used for outcome assessment. The memory ablation evaluates this context jointly while keeping the DP controller unchanged.


\section{Method}
\label{sec:method}
\begin{figure*}[t]
    \centering
    \includegraphics[width=1\linewidth]{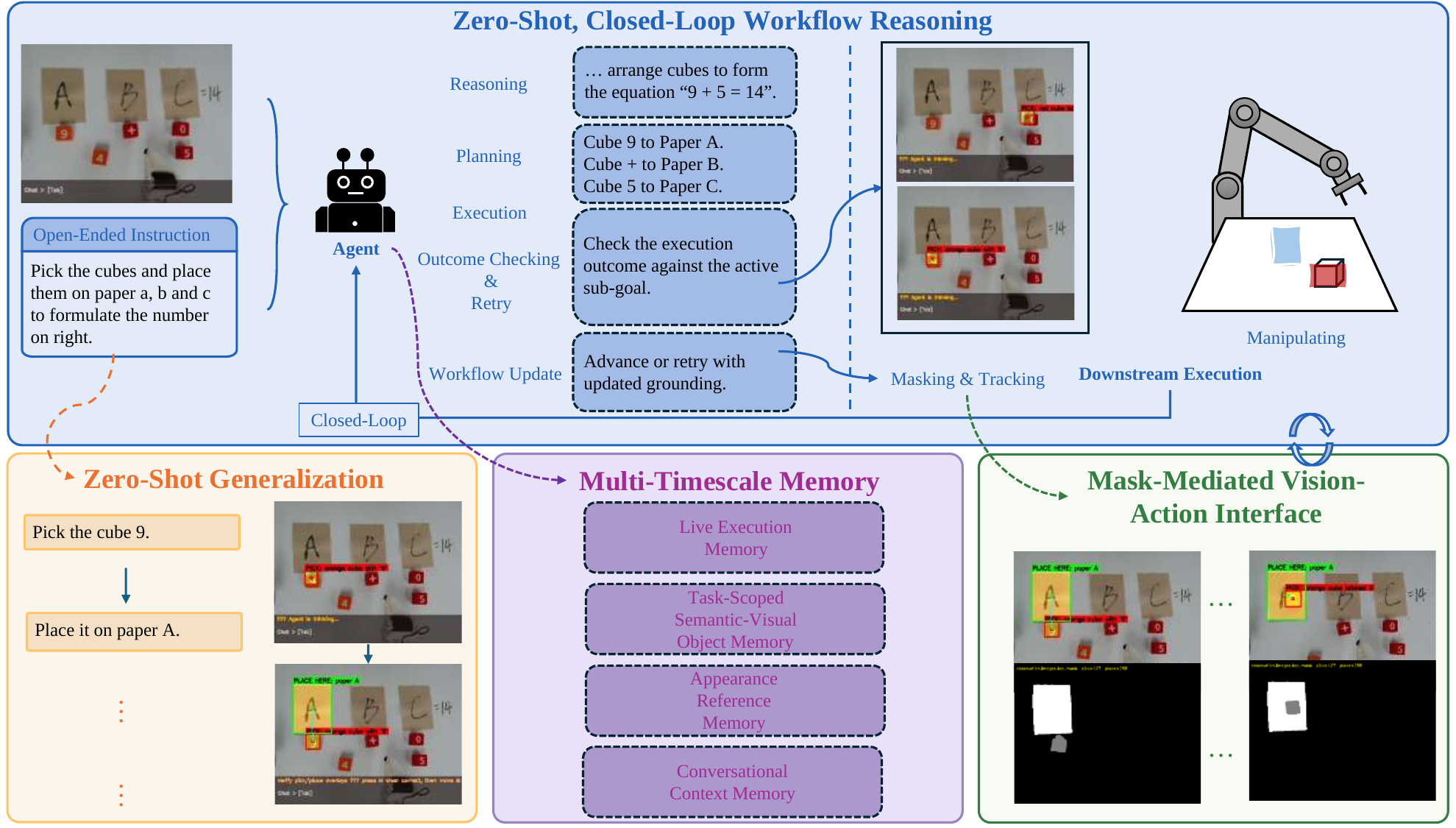} 
     \caption{\textbf{Overview of ACE.} The agent decomposes an instruction into semantic sub-goals, grounds each into a tracked pick-and-place mask, and invokes a reusable VA policy. The scene-state manager retains live execution state and persistent object and task context. Post-execution feedback guides workflow progression, renewed grounding, and retries.}
    \label{fig:framework}
\end{figure*}

We consider tabletop manipulation from a live image stream $I_1, I_2, \dots$ and a natural-language instruction $u$. ACE composes task-level semantic reasoning with a reusable Vision-Action (VA) policy. Let $\mathcal{D}_{\text{prim}}$ denote the generic mask-conditioned pick-and-place demonstrations used to train this policy, and $\mathcal{T}_{\text{eval}}$ the semantic evaluation tasks. \emph{Task-level zero-shot composition} means that the policy is trained on $\mathcal{D}_{\text{prim}}$ without full-task demonstrations from $\mathcal{T}_{\text{eval}}$; the agent generates object selection and task sequencing at inference.

\subsection{Semantic Workflow and Execution Feedback}

Given instruction $u$ and context $c$, comprising the current scene information and retained task context, the agent generates a workflow:
\[
W = \mathcal{P}(u, c) = (w_1, w_2, \dots, w_K),
\]
where each step $w_k$ specifies an object--destination assignment $(o_k,d_k)$. Here, $o_k$ identifies the object selected under the task constraints, and $d_k$ identifies its intended destination. Their current image regions are supplied by grounding.

Each active step retains its semantic assignment as grounding updates its regions. The resulting observation and execution report guide workflow progression or renewed grounding and execution.
\paragraph{External Executable Skills}
ACE exposes two robot-facing skills: \texttt{MaskInterface} grounds the active sub-goal, and \texttt{PickPlace} executes the object-to-destination transfer. The agent framework manages workflow decomposition, memory queries, and sub-goal updates.

\paragraph{Workflow Example}
Figure~\ref{fig:framework} illustrates an instruction to form the number 14. The agent selects $9+5$ and assigns cube 9 to paper A, the addition cube to B, and cube 5 to C. For each transfer, grounding produces the combined pick-and-place mask, the controller executes, and the agent assesses the resulting scene to advance or retry.
\subsection{Mask-Mediated Vision-Action Interface}

For each active workflow step $w_k$, \texttt{MaskInterface} generates a unified pick-and-place mask:
\[
M_k = \mathcal{G}(w_k, I_t, m_t),
\]
where $I_t$ is the current image, $m_t$ denotes the relevant memory state, and $M_k \in \{0,127,255\}^{H \times W}$ is a unified 8-bit grayscale mask. Pixel values encode the pick target, place target, and background:
\[
M_k(x,y) =
\begin{cases}
127, & \text{pick target}, \\
255, & \text{place target}, \\
0, & \text{otherwise}.
\end{cases}
\]

Given $M_k$, ACE initializes a tracker $T_k = \mathcal{T}_{\text{init}}(I_t, M_k)$ and updates it on subsequent frames to obtain $(T_k, M_{k,t+1}) = \mathcal{T}_{\text{update}}(T_k, I_{t+1})$. The tracked mask supplies the VA policy with updated target regions throughout the transfer, while the pixel labels retain their pick and place roles.

\paragraph{Roles and object identity}
The two nonzero labels are scoped to the active transfer. A value of 127 can designate cube 9 in one step and the addition cube in the next; the assignment $(o_k,d_k)$ retains their semantic identities upstream. The mask communicates their current regions and pick/place roles to the controller. Color and printed content enter object selection through the agent and grounding interface, while the controller receives the resulting spatial assignment. This separation allows an instruction to change which object is manipulated without changing the policy's input convention.

\subsection{Reusable Pick-and-Place Control}

\texttt{PickPlace} executes the active sub-goal using masks from \texttt{MaskInterface}. The VA policy conditions on two consecutive observations and returns an action chunk:
\[
\hat{a}_{t:t+L-1} = \pi_{\mathrm{VA}}(q_{t-1:t}, M_{k,t-1:t}),
\]
where $q_t$ is the robot proprioceptive state. DP is configured with a prediction horizon of 64 and $L=32$ action steps per policy call. ACE executes these actions in a receding-horizon manner while updating observations and masks. The controller receives masks and proprioception; upstream modules process instructions and scene images. The same policy is used to transfer number cubes, operator cubes, and retrieval targets. After each attempt, the skill returns an execution report to the workflow manager.

\subsection{Execution State and Persistent Context}
\label{sec:memory}

Planning, tracking, and physical execution proceed on different timescales. ACE coordinates them through a shared scene-state manager with four memory components:
\begin{itemize}
    \item \textbf{Live execution memory} maintains the current workflow position, active masks, tracker state, and execution progress on fast timescales.
    \item \textbf{Task-scoped semantic-visual memory} stores object names, aliases, roles, and visual associations for subsequent grounding and workflow decisions.
    \item \textbf{Appearance reference memory} stores image features or crops associated with known objects as references for later grounding calls.
    \item \textbf{Conversational context memory} maintains a bounded agent-call history containing the initial instruction and previous reasoning and tool invocations.
\end{itemize}
The live state follows tracking and control, while persistent object and conversational context remains available across agent calls. This distinction matters when manipulation changes visible object cues. After a cube rotates, its printed pattern may be hidden; earlier object associations and task context remain available to the agent alongside the current image when it assesses completion.

\subsection{Sub-goal Completion and Retries}
The execution cycle connects the active assignment to the next workflow decision:
\begin{enumerate}
    \item \textbf{Ground the active assignment.} The agent invokes \texttt{MaskInterface} to locate $o_k$ and $d_k$ in the current scene and initialize their tracked mask.
    \item \textbf{Execute the transfer.} \texttt{PickPlace} uses the mask-conditioned policy while tracking updates the target regions. The skill then returns an execution report.
    \item \textbf{Assess the sub-goal.} The agent considers the resulting observation and report together with the active assignment and available context. A completed sub-goal advances the workflow; otherwise, the agent regrounds and retries.
\end{enumerate}
The trial terminates when the agent determines three times that the same active sub-goal has not been completed. The count is attached to that sub-goal. Retries remain within the original trial. GA scores the final localization outcome of each required target, as defined in Section~\ref{sec:evaluation_metrics}.

The memory-free variant retains the execution cycle, active workflow position, and tracker state while removing persistent object and conversational context (Appendix~\ref{sec:ablation_setup}).

\begin{table*}[t]
    \centering
    \caption{\textbf{Main results on multi-step tabletop tasks.}
Each method--task setting uses 20 physical trials. SR denotes task success and FGS the formula-assembly score (maximum 20). GA measures final target-localization accuracy: pick and place are counted separately, and intermediate retries add no extra items.}
    \label{tab:main_results}
    \resizebox{\textwidth}{!}{
    \begin{tabular}{l c c c c c}
        \toprule
        & \multicolumn{3}{c}{\textbf{Semantic Formula Assembly}} & \multicolumn{2}{c}{\textbf{Constraint Retrieval}} \\
        \cmidrule(lr){2-4} \cmidrule(lr){5-6}
        \textbf{Method} & \textbf{avg. FGS (Max 20)} & \textbf{SR (\%)} & \textbf{GA (\%)}  & \textbf{SR (\%)} & \textbf{GA (\%)} \\
        \midrule
        ACT (Imitation) & 3.2 & 0.0 & -- & 0.0 & -- \\
        $\pi_{0.5}$ (End-to-End VLA) & 3.0 & 0.0 & -- & 0.0 & -- \\
        \midrule
        ACE (w/o persistent memory) & 16.8 & 55.0 & 96.7 & 70.0 & 90.0 \\
\textbf{ACE (Full Framework)} & \textbf{18.0} & \textbf{70.0} & \textbf{100.0} & \textbf{80.0} & \textbf{97.5} \\
        \bottomrule
    \end{tabular}
    }
\end{table*}

\section{Experiments}
\label{sec:experiments}

We evaluate whether a controller trained on approximately one hour of generic pick-and-place demonstrations can execute semantic tasks without full-task demonstrations of either evaluation task. The main results assess this task-level zero-shot composition capability. Two matched ablations then examine the controller's visual input and the agent's persistent context, measuring their contributions within the deployed framework.

\subsection{Physical Setup and Protocol}
\label{exp}
We deploy the agentic reasoning engine on a cloud server with four NVIDIA RTX 3090 GPUs. The mask interface and VA policy run locally on a single RTX 3090. Each setting in Tables~\ref{tab:main_results} and~\ref{tab:ablation_results} is evaluated over 20 independent physical trials. We randomize initial object layouts and paraphrase instructions while preserving their semantic meaning. Implementation details are provided in Appendix~\ref{Implementation}.

All trials in Tables~\ref{tab:main_results} and~\ref{tab:ablation_results} run autonomously after the initial instruction. Any grounding updates and subsequent execution attempts are handled by the evaluated system. For agent-controlled trials, the trial terminates when the agent determines three times that the same active sub-goal has not been completed.

\subsection{Tasks}

We evaluate ACE on two multi-step tabletop tasks requiring instruction-dependent object selection and execution. These tasks differ in semantic constraints, object selection logic, and workflow structure while sharing the same reusable pick-and-place primitive:
\textbf{Semantic Formula Assembly:} The system selects number and operator cubes to form an expression satisfying the requested arithmetic condition. In all trials, three visible paper regions labeled A, B, and C specify the destinations for the first operand, operator, and second operand, respectively.
\textbf{Constraint Retrieval:} The system must evaluate logical or numerical constraints to identify specific target objects and transport them to a designated spatial region.
Examples are shown in Appendix~\ref{real}.

\subsection{Evaluation Metrics}
\label{sec:evaluation_metrics}
We report task Success Rate (SR), Grounding Accuracy (GA), and a Fine-Grained Score (FGS) for Semantic Formula Assembly. SR measures complete task success. GA measures the fraction of required targets that are correctly localized in the final grounding result, counting pick and place separately:
\[
\mathrm{GA}\;(\%) = 100\,\frac{N_{\mathrm{correct}}}{N_{\mathrm{targets}}},
\]
where $N_{\mathrm{targets}}$ is the total number of required targets and $N_{\mathrm{correct}}$ is the number correctly localized in the final result. Each required target is counted once; intermediate attempts and retries do not add items. FGS measures partial completion through semantic selection, grasping, placement, and final equation correctness, with a maximum of 20 points. The scoring rules are provided in Appendix~\ref{Metrics}.

\subsection{Baselines}
We compare ACE with ACT~\cite{zhao2023learningfinegrainedbimanualmanipulation} and $\pi_{0.5}$~\cite{intelligence2025pi05visionlanguageactionmodelopenworld}, trained or fine-tuned with approximately one hour of full-task demonstrations collected using LeRobot~\cite{cadene2026lerobot}. These demonstrations include complete formula-assembly and constraint-retrieval trajectories. ACT is trained separately for each evaluation task. ACE's VA policy uses approximately one hour of generic mask-conditioned pick-and-place demonstrations. This protocol evaluates whether a primitive-trained controller combined with task reasoning can execute the semantic tasks alongside models trained on complete task trajectories.

We also train a separate standalone Diffusion Policy for each task using the corresponding full-task demonstrations used for ACT (Table~\ref{tab:ablation_results}(b)).



\begin{table*}[t]
    \centering
    \caption{\textbf{Visual-input ablation and controller comparisons.}
Each setting uses 20 physical trials. Top: single-cube grasp success. The two DP input configurations use the same test objects and matched training and execution settings. Bottom: task performance using full-task demonstrations for standalone DP and primitive-level demonstrations for ACE.}
    \label{tab:ablation_results}
    \resizebox{\textwidth}{!}{
    \begin{tabular}{l c c c c c}
        \toprule
        \multicolumn{6}{c}{\textbf{(a) Single-Cube Grasp Results}} \\
\midrule
\multicolumn{2}{c}{\textbf{Method}} &
\multicolumn{2}{c}{\textbf{Setting}} &
\multicolumn{2}{c}{\textbf{Grasp SR (\%)}} \\
\cmidrule(lr){1-2} \cmidrule(lr){3-4} \cmidrule(lr){5-6}
\multicolumn{2}{c}{$\pi_{0.5}$ LoRA} &
\multicolumn{2}{c}{Seen} &
\multicolumn{2}{c}{60.0} \\
\multicolumn{2}{c}{$\pi_{0.5}$ LoRA} &
\multicolumn{2}{c}{Seen + Unseen} &
\multicolumn{2}{c}{40.0} \\
\multicolumn{2}{c}{DP (Mask + RGB)} &
\multicolumn{2}{c}{Shared test set} &
\multicolumn{2}{c}{30.0} \\
\multicolumn{2}{c}{ACE / DP (Mask Only)} &
\multicolumn{2}{c}{Shared test set} &
\multicolumn{2}{c}{90.0} \\
        
        \midrule
        \midrule
        \multicolumn{6}{c}{\textbf{(b) End-to-End Task Comparison}} \\
        \midrule
        & \multicolumn{3}{c}{\textbf{Semantic Formula Assembly}} & \multicolumn{2}{c}{\textbf{Constraint Retrieval}} \\
        \cmidrule(lr){2-4} \cmidrule(lr){5-6}
        \textbf{Method} & \textbf{avg. FGS (Max 20)} & \textbf{SR (\%)} & \textbf{GA (\%)}  & \textbf{SR (\%)} & \textbf{GA (\%)} \\
        \midrule
        Standalone DP & 3.4 & 0.0 & -- & 0.0 & -- \\
        \textbf{ACE (Full Framework)} & \textbf{18.0} & \textbf{70.0} & \textbf{100.0} & \textbf{80.0} & \textbf{97.5} \\
        \bottomrule
    \end{tabular}
    }
\end{table*}

\section{Results}
\label{subsec:results}

Table~\ref{tab:main_results} reports task completion, partial progress, and target grounding. ACE completes 14 of 20 Semantic Formula Assembly trials (70.0\% SR), with an average FGS of 18.0 out of 20, and 16 of 20 Constraint Retrieval trials (80.0\% SR). Formula success requires a semantically valid expression and correct placement of its operands and operator; retrieval success requires selecting and placing objects that satisfy the instruction. Both are executed with the same controller trained on generic transfers, with complete evaluation-task trajectories absent from its training data.

ACT and $\pi_{0.5}$ complete no full tasks under the training configurations in Section~\ref{sec:experiments}, with formula-assembly FGS values of 3.2 and 3.0, respectively. The standalone DP comparison in Table~\ref{tab:ablation_results}(b) also records no complete tasks and an FGS of 3.4. 

ACE achieves 100.0\% GA on Semantic Formula Assembly and 97.5\% on Constraint Retrieval, measuring final target-localization outcomes. The difference between 100.0\% GA and 70.0\% task success in formula assembly shows why grounding and complete execution are evaluated separately. FGS further records progress through selection, grasping, placement, and equation formation.


\section{Ablations and Controller Comparisons}

\paragraph{Persistent memory}
In Table~\ref{tab:main_results}, the full ACE framework achieves 70.0\% SR and 18.0 FGS on Semantic Formula Assembly, compared with 55.0\% SR and 16.8 FGS without persistent memory. On Constraint Retrieval, the corresponding success rates are 80.0\% and 70.0\%. The observed gains are therefore 15 and 10 percentage points, corresponding to three and two additional successful trials. Grounding accuracy also increases from 96.7\% to 100.0\% on Semantic Formula Assembly and from 90.0\% to 97.5\% on Constraint Retrieval. Both variants use the same DP controller. The ablation removes persistent conversational and object context while retaining the live workflow and tracking state (Appendix~\ref{sec:ablation_setup}).

\begin{figure*}[t]
    \centering
    \includegraphics[width=1\linewidth]{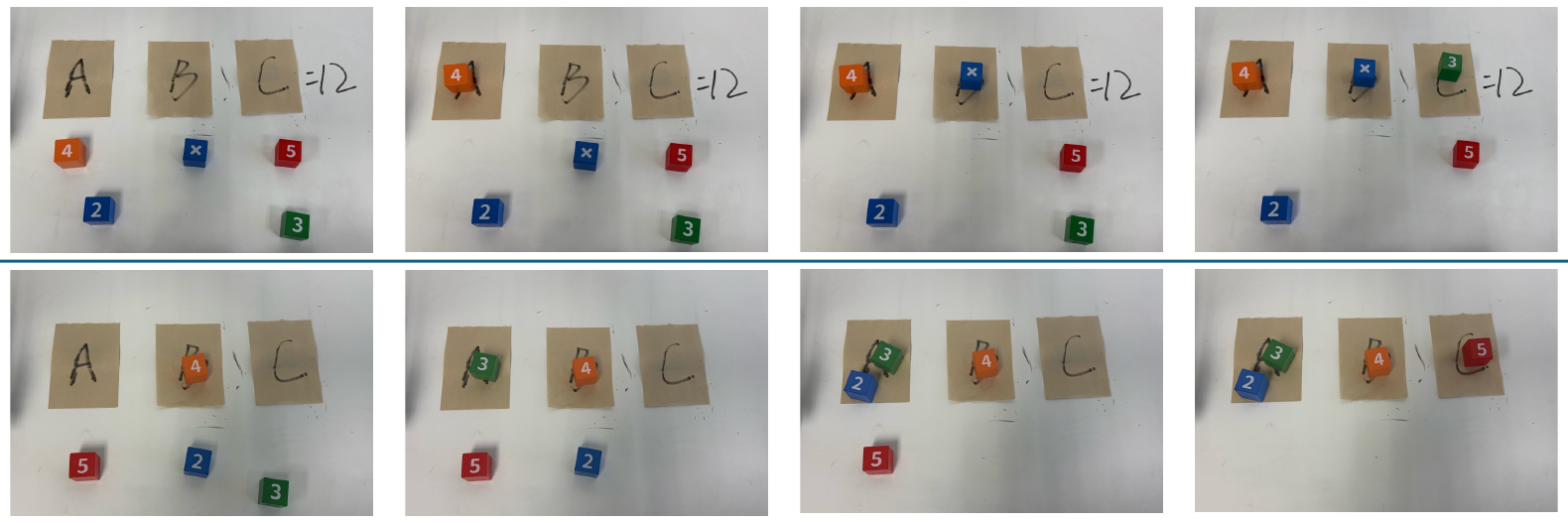} 
     \caption{\textbf{Execution sequences for the evaluation tasks.} \textbf{Top:} Semantic Formula Assembly arranges cubes into a valid expression. \textbf{Bottom:} Constraint Retrieval selects and places cubes according to the requested conditions (e.g., values $<4$ to the left and $>4$ to the right). }
    \label{fig:real}
\end{figure*}

\paragraph{Completion uncertainty after appearance changes}
In observed memory-free trials, cube rotation could hide the identifying surface pattern, leaving the agent uncertain whether it had completed the task. This is an ambiguity in interpreting the post-action scene: the current appearance may no longer expose the cue used to identify the object earlier. Persistent memory retains those earlier object associations and task information for subsequent assessment. This qualitative observation illustrates a role for context alongside the success-rate differences above.

\paragraph{Visual-input ablation}
The two DP configurations in Table~\ref{tab:ablation_results}(a) use identical training and execution settings and are evaluated on the same objects over 20 trials each. Mask-only DP succeeds in 18 of 20 trials (90.0\%), compared with 6 of 20 (30.0\%) for Mask + RGB, an observed difference of 60 percentage points. This comparison supports the mask-only visual interface under the evaluated conditions. Unseen objects in the evaluation have out-of-distribution colors and surface patterns.

The grasping comparison evaluates the controller's spatial conditioning under the common training-data regime. The memory ablation evaluates context reuse during complete semantic tasks while holding the controller fixed. These component comparisons support the design choices used in the main task-level zero-shot evaluation.

The $\pi_{0.5}$ LoRA controller, fine-tuned on 300 grasp demonstrations, achieves 60.0\% grasp success on seen objects and 40.0\% on the mixed seen/unseen setting. The standalone DP results in Table~\ref{tab:ablation_results}(b) follow the full-task setup in Section~\ref{sec:experiments}.


\section{Limitations and Future Work}
\label{limitation}
The evaluation covers two cube-manipulation tasks and one pick-and-place primitive. Persistent object and conversational context are ablated jointly. Recovery observations are qualitative; verification accuracy, retry success, and execution time are not separately measured.

ACE's design offers a route to broader task goals and scenes. The agent maps semantic requirements to object--destination assignments, while tracked masks provide a shared spatial interface for executing them. This separation could support new selection rules and placement goals within the controller's manipulation capabilities. Future work will evaluate transfer across diverse objects, scene layouts, and task constraints, measure grounding and execution robustness, and add learned primitives beyond pick-and-place.


\section{Conclusion}
\label{sec:conclusion}

ACE composes semantic tasks through object--destination assignments, tracked pick/place masks, and a reusable learned policy. Persistent context supplies earlier object and task information for outcome assessment as manipulation changes the scene. Trained only on generic transfers and without full-task demonstrations, the controller enables ACE to achieve 70\% success on Semantic Formula Assembly and 80\% on Constraint Retrieval.

\useRomanappendicesfalse
\appendices

\section{Task Descriptions}
\label{real}
Figure~\ref{fig:real} shows representative executions of the two evaluation tasks.

\paragraph{Semantic Formula Assembly}
For a requested value $y$, let $o_a$ and $o_b$ denote selected number cubes and $o_{\circ}$ the selected operator cube. A valid selection satisfies
\[
\operatorname{value}(o_a)\;\circ\;\operatorname{value}(o_b)=y.
\]
The corresponding transfers can be written as
\[
W_{\mathrm{formula}}=((o_a,A),(o_{\circ},B),(o_b,C)).
\]
Grounding supplies the current regions for these role-specific transfers. The task combines semantic correctness of the expression with physical placement in the requested order.

\paragraph{Constraint Retrieval}
Let $\mathcal{O}$ denote the objects under consideration and $\phi_u(o)$ the selection condition expressed by instruction $u$. The selected set is $\mathcal{O}_u=\{o\in\mathcal{O}:\phi_u(o)=1\}$. Each selected object is assigned a destination $d_u(o)$ according to the instruction, producing transfers $(o,d_u(o))$. For the example in Fig.~\ref{fig:real}, a cube with value below 4 is assigned to the left region, and one with value above 4 to the right. The agent generates the sequence of transfers; all use the same mask encoding and learned pick-and-place controller as formula assembly.

\section{Implementation Details}
\label{Implementation}
The implementation combines an agent, visual grounding and tracking, and a robot controller.

\textbf{Planner and Workflow Manager:} We implement the core language agent using the Qwen-Agent framework~\cite{qwen-agent-2405}, powered by Qwen3.8-27B~\cite{qwen38}. Qwen processes user instructions and rendered scene context to generate workflow steps and select subsequent skill calls. The two robot-facing skills, \texttt{MaskInterface} and \texttt{PickPlace}, are registered as callable tools within Qwen-Agent. The framework also supports state queries and memory management.

\textbf{Mask-Mediated Vision-Action Interface:} For the mask-mediated grounding ($\mathcal{G}$) and persistent tracking ($\mathcal{T}$), we utilize SAM3~\cite{carion2026sam3segmentconcepts} to generate pick-and-place masks and use Cutie~\cite{cheng2024puttingobjectvideoobject} to update the masks over subsequent frames. 

\textbf{Robot Control:} We use a SO-101 arm~\cite{SO-101} with LeRobot~\cite{cadene2026lerobot} Diffusion Policy~\cite{chi2024diffusionpolicyvisuomotorpolicy}, trained on approximately one hour of generic object-to-region transfers specified by pick/place masks.

\textbf{DP Training Configuration:} The grayscale mask is replicated across three channels to form a $3\times480\times640$ policy input. The policy also receives a six-dimensional robot state and outputs six-dimensional actions. Its ResNet-18 visual backbone uses ImageNet-pretrained weights and 32 spatial-softmax keypoints. The diffusion denoiser uses channel widths $[512,1024,2048]$, kernel size 5, eight groups, and a 128-dimensional diffusion-step embedding. DDPM uses 100 training diffusion timesteps, the \texttt{squaredcos\_cap\_v2} beta schedule, and noise prediction. The configured training budget is 200{,}000 updates with batch size 16 and seed 1000. Adam uses learning rate $10^{-4}$, weight decay $10^{-6}$, $\beta=(0.95,0.999)$, $\epsilon=10^{-8}$, and gradient-norm clipping at 10. A cosine learning-rate schedule follows 500 warmup updates. Visual inputs use mean--standard-deviation normalization; states and actions use min--max normalization. Dataset image transforms are disabled.

\subsection{Fine-grained Scoring}
\label{Metrics}
The Fine-Grained Score (FGS) measures partial completion in Semantic Formula Assembly.

Each episode consists of three sequential pick-and-place steps (first operand, operator, second operand). Each step is assigned a maximum score of 6 points:
\begin{itemize}
    \item \textbf{2 points} for selecting a semantically correct cube. For policies without explicit object-identity outputs, selection is inferred from the cube toward which the robot directs its grasp attempt. A failed grasp can receive selection credit only if the attempted target is semantically correct;
    \item \textbf{2 points} for successfully grasping the selected cube (Physical execution);
    \item \textbf{2 points} for accurately placing the block into the target slot (Placement accuracy).
\end{itemize}

We additionally assign 2 points for final equation correctness, which evaluates whether the assembled expression mathematically matches the target value. The maximum score per episode is:
\[S_{\max} = 3 \times (2 + 2 + 2) + 2 = 20.\]

\paragraph{Grounding counts and task outcomes}
GA counts each required pick or place target once using its final localization outcome. A pair with a correct pick target and an incorrect place target contributes one correct target out of two. A target corrected during retries contributes one correct item; intermediate attempts add no items.

Correct grounding can coexist with failed execution. SR counts completed trials including retries; FGS records partial progress.

\subsection{Ablation and Comparison Settings}
\label{sec:ablation_setup}
The following settings define the memory ablation and additional controller and system comparisons:

\begin{itemize}
    \item \textbf{$\pi_{0.5}$ LoRA single-cube grasping:}
We fine-tune $\pi_{0.5}$ with LoRA on 300 single-cube grasp demonstrations and evaluate grasp success on seen and mixed seen/unseen object settings, with 20 physical trials per setting. Unseen refers to cubes with out-of-distribution colors and surface patterns.
    \item \textbf{DP visual-input ablation:} Mask + RGB receives RGB observations alongside the pick-and-place masks provided by ACE; Mask Only uses the masks as its visual input. The two configurations use identical training and execution settings and the same test objects, with 20 physical trials per configuration.
    \item \textbf{Standalone DP:} We train a separate standalone Diffusion Policy for each evaluation task on the corresponding full-task demonstrations used for ACT.
    \item \textbf{ACE (w/o persistent memory):}
We retain the same DP controller and disable access to the persistent memory described in Section~\ref{sec:memory}: LLM conversation history, object names and aliases, textual object facts, visual identity features, and Code-as-Policies session history.
Object memory is no longer injected into subsequent planning or visual prompts, and the \texttt{record\_task\_object\_knowledge} tool is removed.
The current workflow, step index, bounding boxes, and SAM/Cutie tracking state are retained to support ongoing task execution.
\end{itemize}

\bibliographystyle{IEEEtran}
\bibliography{example} 

\end{document}